\documentclass[letterpaper, 10 pt, conference]{ieeeconf}  

\IEEEoverridecommandlockouts                              

\title{\LARGE \bf
Human Observation-Inspired Trajectory Prediction for Autonomous Driving in Mixed-Autonomy Traffic Environments
}

\author{Haicheng Liao$^{*1}$, Shangqian Liu$^{*1}$, Yongkang Li$^{2}$, Zhenning Li$^{\dag3}$,  Chengyue Wang$^{4}$, \\Yunjian Li$^{5}$, Shengbo Eben Li$^{6}$, and Chengzhong Xu$^{1}$
\thanks{*Both authors contributed equally to this research.}
\thanks{\dag Corresponding author.}
\thanks{$^{1}$State Key Laboratory of Internet of Things for Smart City and Department of Computer and Information Science, University of Macau, Macau SAR, China. Email: 
        \{\tt\small yc27979, mc25671, czxu\}@um.edu.mo}%
\thanks{$^{2}${Department of Information and Software Engineering, University of Electronic Science and Technology of China, Chengdu, China.} Email: {\tt\small franklin1234560@163.com}}%
\thanks{$^{3}${State Key Laboratory of Internet of Things for Smart City and Departments of Civil and Environmental Engineering and Computer and Information Science, University of Macau, Macau SAR,  China.} Email: {\tt\small zhenningli@um.edu.mo}}
\thanks{$^{4}${State Key Laboratory of Internet of Things for Smart City and Departments of Civil and Environmental Engineering, University of Macau, Macau SAR,  China.} Email: {\tt\small emailcyw@gmail.com}}
\thanks{$^{5}${Faculty of Innovation Engineering Macau University of Science and Technology, Macau SAR, China.} Email: {\tt\small liyunjian@must.edu.mo}}
\thanks{$^{6}${School of Vehicle and Mobility, Tsinghua University, Beijing, China.} Email: {\tt\small lishbo@tsinghua.edu.cn}}
}

\usepackage{amsmath}
\usepackage{amssymb}
\usepackage{graphicx}
\usepackage{multirow}
\usepackage{graphicx}
\usepackage{textcomp}
\usepackage{subfigure}
\usepackage{url}
\usepackage{verbatim}
\usepackage{graphicx}
\usepackage{amssymb}
\usepackage{amsmath}
\usepackage{array,booktabs}
\usepackage{multirow}
\usepackage{bm}
\usepackage{bbding}
\usepackage{pifont}
\usepackage{makecell}
\usepackage{array}
\usepackage{tabularx}
\usepackage{newfloat}
\usepackage{listings}
\usepackage{algorithm}
\usepackage{algorithmic}
\usepackage{hyperref}
\usepackage{xcolor}
\begin{document}

\maketitle
\thispagestyle{empty}
\pagestyle{empty}

\begin{abstract}
In the burgeoning field of autonomous vehicles (AVs), trajectory prediction remains a formidable challenge, especially in mixed autonomy environments. Traditional approaches often rely on computational methods such as time-series analysis. Our research diverges significantly by adopting an interdisciplinary approach that integrates principles of human cognition and observational behavior into trajectory prediction models for AVs. We introduce a novel ``adaptive visual sector" mechanism that mimics the dynamic allocation of attention human drivers exhibit based on factors like spatial orientation, proximity, and driving speed. Additionally, we develop a ``dynamic traffic graph" using Convolutional Neural Networks (CNN) and Graph Attention Networks (GAT) to capture spatio-temporal dependencies among agents. Benchmark tests on the NGSIM, HighD, and MoCAD datasets reveal that our model (GAVA) outperforms state-of-the-art baselines by at least 15.2\%, 19.4\%, and 12.0\%, respectively. Our findings underscore the potential of leveraging human cognition principles to enhance the proficiency and adaptability of trajectory prediction algorithms in AVs. The code for the proposed model is available at our \hypersetup{hidelinks}\href{https://github.com/Petrichor625/Gava.git}{\color{purple}{Github}}.
\end{abstract}

\section{INTRODUCTION}
In the pursuit of developing fully autonomous vehicles (AVs), one of the most daunting challenges is to accurately predict vehicle trajectories with the same level of proficiency as human drivers, especially in a mixed autonomy environment where AVs coexist with human-driven vehicles [1]. While existing research largely focuses on computational methods grounded in time-series analysis, our work takes a distinct direction. We adopt an interdisciplinary approach that seeks to integrate principles of human cognition and observational behavior.

To appreciate the relevance of this interdisciplinary approach, it's essential to understand the complexities of human driving behavior. These complexities are deeply rooted in a synergistic interplay between observational faculties and cognitive decision-making processes. It is worth noting that empirical evidence suggests that human drivers rely on visual information for approximately 90\% of their driving decisions [2]. Given this, we are led to a pivotal research question: Can a trajectory prediction model for AVs be formulated that incorporates insights from the visual perception and attention allocation strategies exhibited by human drivers?

Answering this question necessitates an examination of the intrinsic limitations of human cognitive capacity and their measurable impacts on driving behavior. Research indicates that the human brain can effectively process information related to only a limited number of external agents, usually not exceeding four [3]. This limitation sets the stage for the priority schema that human drivers subconsciously follow. Specifically, when faced with multiple stimuli, drivers allocate attention based on factors like spatial orientation and proximity to the vehicle [4]. Additionally, given the increased risks of frontal collisions, a significant portion of cognitive resources is channeled towards the \textbf{central visual field} [5]. This focus enables quicker detection of and reaction to potential obstacles lying ahead [6].
\begin{figure}[t]
  \centering  \includegraphics[width=\linewidth]{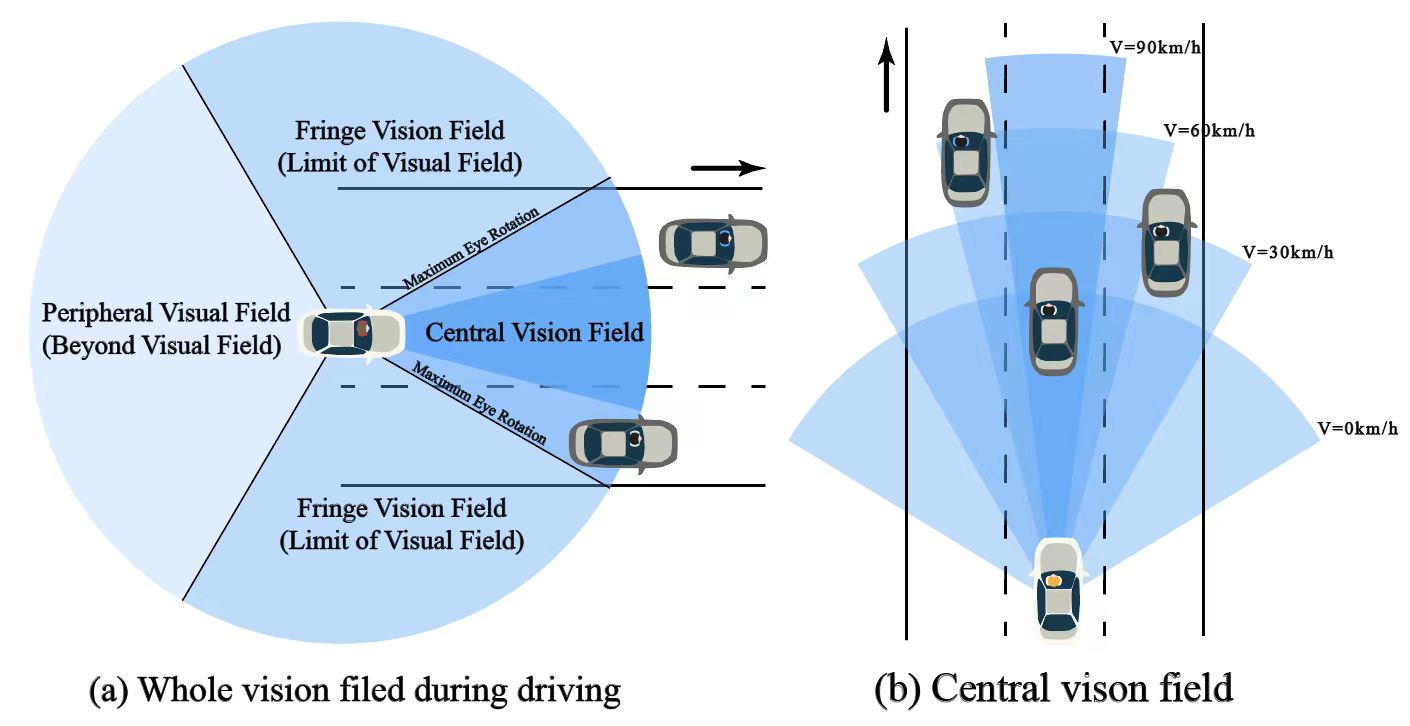} 
  \caption{Adaptive dynamic mechanism of the human driver's attention. \textbf{Left}: The human driver's visual field can be roughly divided into three parts: \textbf{central}, \textbf{fringe}, and \textbf{peripheral} vision fields. The central vision field receives the most attention, while visual information from the fringe and peripheral vision fields is prioritized during maneuvering changes, which are observed through side and rearview mirrors and receive comparatively less attention and observation time. \textbf{Right}: The coverage angle of the central vision field adjusts rapidly with velocity in real-time.}
  \label{fig1}
\end{figure}
In our study, we also pay close attention to a nuanced, yet vital, aspect of human driving behavior: the adaptability of the driver's 'visual sector,' which is defined by both its radius and angle. Notably, this sector isn't static but dynamically adapts in real-time in relation to driving speed [7]. Counterintuitively, this adaptability manifests in a narrower visual sector at higher speeds and a wider one at lower speeds. The reason for this lies in human cognitive and perceptual adjustments. At higher speeds, the narrowed focus allows the driver to concentrate on the road directly ahead, ensuring a quicker response to sudden obstacles or changes. On the other hand, a broader focus at lower speeds enables the driver to absorb more peripheral information, such as roadside activity, pedestrians, and traffic signals, which becomes more relevant at slower paces. By incorporating an adaptive visual sector that adjusts its field-of-view based on speed, our model mimics this inherently human trait, providing a more intuitive and context-sensitive trajectory prediction.

Building on these insights, our model aims for a more human-like, adaptive trajectory prediction algorithm for AVs. The central contributions of our research are threefold:
 \begin{itemize}
    \item We introduce a sophisticated pooling mechanism that replicates human attention allocation with a novel \textbf{adaptive visual sector}. This mechanism dynamically adapts its focus in real-time, allowing the model to capture important perceptual cues from different scenes.
    
    \item We introduce a novel \textbf{dynamic traffic graph} to extract the interaction of agents using a unique \textbf{topology graph structure} constructed using Convolutional Neural Networks (CNN) and Graph Attention Networks (GAT). Enhanced by multi-head attention mechanisms, this architecture provides a robust means to model spatio-temporal dependencies and generate multiple hypothesis trajectories with corresponding credibility.
    
    \item In benchmark tests on the NGSIM, HighD, and MoCAD datasets, our model outperforms the state-of-the-art (SOTA) baselines by at least 15.2\%, 19.4\% and 12.0\%, respectively, demonstrating its impressive accuracy and applicability in various traffic scenarios, including \textbf{highways} and \textbf{dense urban areas}.
\end{itemize}

\section{RELATED WORKS}
\textbf{Deep Learning-based trajectory prediction.} The rise of deep learning technologies has significantly propelled the field of trajectory prediction in autonomous vehicles (AVs). Long Short-Term Memory (LSTM) networks are renowned for their ability to model nonlinear time-series data, serving as a fundamental component in various trajectory prediction studies [9,10,11]. LSTMs have been particularly instrumental in predicting both longitudinal and lateral vehicle trajectories on highways [12]. Convolutional Neural Networks (CNNs) have also seen application in this domain, especially in modeling social interactions between vehicles through techniques like Social Pooling [13,14]. Hybrid CNN-LSTM models have also demonstrated promising results [15]. Further enriching the field are Graph Neural Networks (GNNs), which have been employed in Generative Adversarial Network (GAN)-based frameworks [16] and lightweight Graph Convolutional Network (GCN)-based models [17]. The introduction of multi-head attention mechanisms has led to advanced models such as HDGT, which combines Transformers [40] with Recurrent Neural Networks (RNNs) [18], and the spatio-temporal model STDAN [19] and BAT [41], known for its high prediction accuracy.

\textbf{Vision Aware Mechanisms in Traffic Behavior.} One body of research explores the concept of the Visual Field, highlighting that a driver's field of view changes based on vehicle speed[6]. Specifically, slower speeds result in a broader focus on the surrounding environment, while higher speeds narrow the focus, emphasizing more distant views [20]. Another study builds upon this by showing that drivers can recognize more details and have a wider visual perspective at lower speeds. This research also delves into the implications of such visual mechanisms for accident rates [21]. A more recent study introduces the term "visual attention" to encompass these phenomena. Using eye-tracking technology, the study finds that drivers' focus shifts based on varying speeds and road types. The focus is generally divided into looking ahead for vehicle control and looking to the sides for lane-changing decisions, with these focus areas dynamically adapting based on the speed of the vehicle [22].

\section{PROBLEM FORMULATION}
This study aims to forecast the future trajectory of a target vehicle over a fixed time horizon $T$ by leveraging historical states of both the target vehicle and its surrounding agents over a time duration of $f$.

We define $x_t^i$ as the state of the target vehicle (superscript 0) and all observed surrounding vehicles (superscript 1 to N) at a given time $t$. This state includes position coordinates, velocity, acceleration, vehicle type, lateral and longitudinal behavior, and lane information.

At each time step $t$, given the the historical states of the spans $T$, 
\begin{equation}
    \bm{X}=\{x_{t-T:t}^{0}, x_{t-T:t}^{1}, x_{t-T:t}^{2}, \ldots, x_{t-T:t}^{N}\}
\end{equation}, 
which used for the inputs of the model. 

The primary objective is to predict the trajectory of the target vehicle, where prediction spans from time step $ T+1 $ to $ T+F $:
\begin{equation}
    \bm{Y} = \{ y_{T+1}^{0}, y_{T+2}^{0}, \ldots, y_{T+F}^{0} \}, 
\end{equation}
 where $ F $ is the prediction horizon. Each $ y_{T+f}^{0} $ (where $ 1 \leq f \leq F $) represents the future $ x $ and $ y $ coordinates of the target vehicle at time $ T+f $.

In our model, the distribution of $ y_{T+f}^{0} $ is captured using a bivariate Gaussian distribution. This is denoted as $ P(\bm{Y}|\bm{X}) $, which quantifies the uncertainty associated with each predicted trajectory point.

\section{PORPOSED MODEL}
Fig. 2 shows the architecture of our model,  including Context-aware, Interaction-aware, Vision-aware, and Priority-aware modules.
Collectively, these modules are designed to emulate the human observation process during driving.

\begin{figure*}[t]
  \centering  \includegraphics[width=0.95\linewidth]{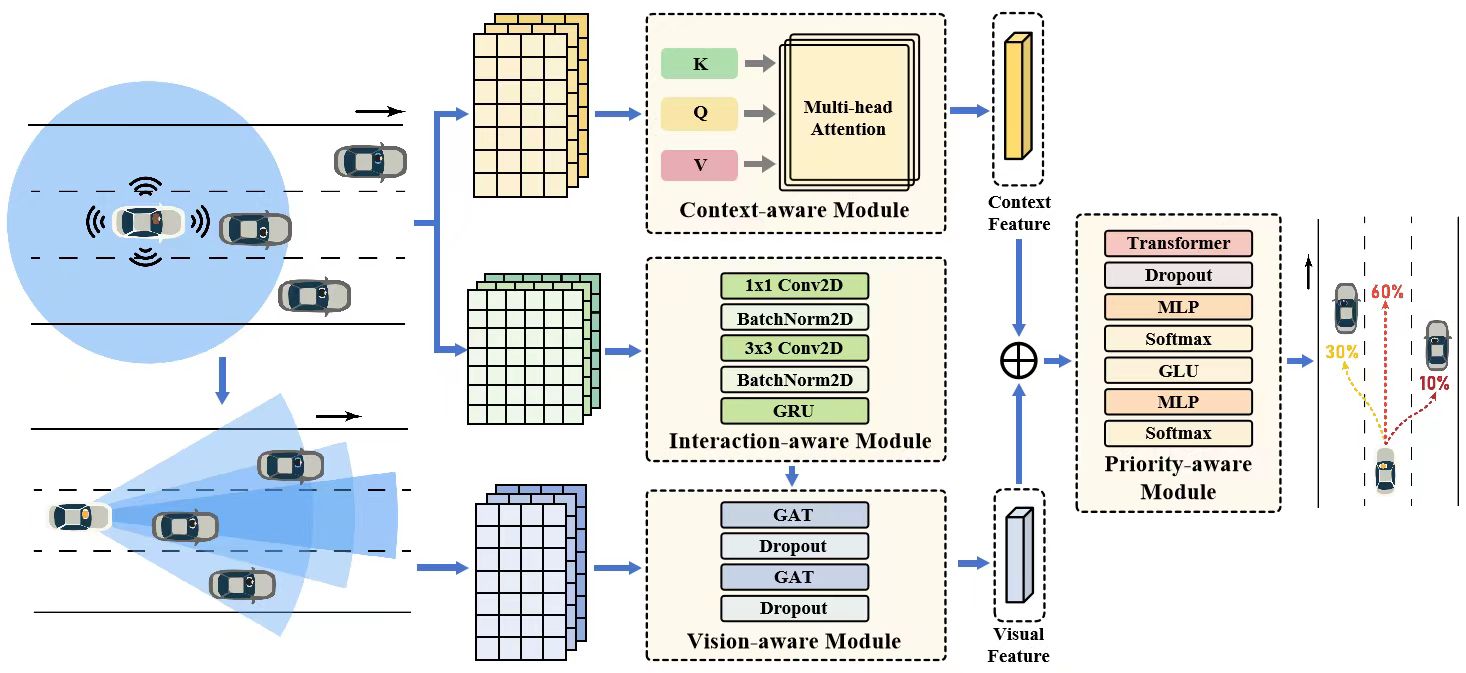} 
  \caption{A general overview of Gava. The original data is processed through the Context-Aware Module to extract high-dimensional temporal information, resulting in the formation of the context feature. Meanwhile, raw data is processed through the Visual-Aware and Interaction-Aware Modules, yielding Visual Features. Finally, the Priority-Aware Module is employed to generate multi-modal trajectory predictions based on the fusion of the two features.}
  \label{fig2}
\end{figure*}

\subsection{Context-aware Module}
The Context-aware Module aims to efficiently capture temporal data by amalgamating the historical states of the target vehicle and its observed surrounding agents [19]. Specifically, a Gated Recurrent Unit (GRU) is used to process temporal data for both the target and the surrounding agents, encoding their trajectories over each observation interval and allowing the data to be amalgamated. A fully connected layer transforms the input motion state $x_{t}^{i}$ into an embedded representation. This transformation uses a Multi-Layer Perceptron (MLP) with learnable parameters, activated by an Exponential Linear Unit (ELU):
\begin{equation}
\resizebox{\linewidth}{!}{
$\operatorname{ELU}(x_{t}^{i})= \begin{cases}(x_{t}^{i}, W_{\text{E}})_\text{MLP}, & \text{ if }(x_{t}^{i}, W_{\text{E}})_\text{MLP} \geq 0 \\ \alpha(e^{(x_{t}^{i}, W_{\text{E}})}-1)_\text{MLP}, & \text{ if }(x_{t}^{i}, W_{\text{E}})_\text{MLP} < 0\end{cases}$
}.
\end{equation}

Then GRU cells are used to encode each $x_{i}^{t}$ for each vehicle $i$,
$h_{t}^{i}=\operatorname{GRU}\left(x_{t}^{i}, h_{t-1}^{i}, W_{G}\right),$
where $W_{G}$ are the learnable parameters common to all vehicles. The final GRU state for each vehicle can be expected to encode the motion state of that vehicle.
Formally,
\begin{equation}
\alpha_{t} = \text{softmax}\left(\frac{\left\langle q_{t}, K_{t}\right\rangle}{\sqrt{d_k}}\right)
\end{equation} 
where the $\alpha$ is computed by taking the dot product of the query($q$) and key($K$) vectors and applying a SoftMax operation. The features of the neighbor vehicles are aggregated using the equation:
\begin{equation}Head_t = \sum_{i=1}^{N} \alpha_{t}^{i} v_{t}^{i},\end{equation}
where the weighted sum is then computed using the value ($v$). Next, Gated Linear Units (GLU) are used to extract social dependencies and features from the target vehicle, resulting in $Head_t$. All multi-head information is processed and combined into the context feature: ${H}^{0}=\left[{h}{1}^{0}, {h}{2}^{0}, \cdots, {h}_{T}^{0}\right]$. This output is a collection of historical information processed by the attention mechanism. It expresses the mapping of temporal information from neighboring vehicles to the temporal information of the target vehicle.

\subsection{Interaction-aware Module}
The Interaction-aware Module focuses on capturing spatial interactions between vehicles [39]. This is achieved by aggregating spatial data through a graph polymerization operation. Specifically, three matrices, denoted as $A^{vel}$, $A^{acc}$, and $A^{man}$, are used to represent the variations in velocity, acceleration, and behavioral consistency among vehicles within the same observation area at time $t$. These matrices are then merged into a 3D tensor ${D}$:
\begin{equation}
{D} = \{A^{vel}, A^{acc}, A^{man}\}
\end{equation}
A 2D convolutional layer with a kernel size of $1 \times 1$ is applied to expand the number of channels in the tensor. Another convolutional layer with a kernel size of $3 \times 3$ is then applied to extract channel-specific features. Batch normalization and dropout techniques are used to improve the performance and generalization of the model. The processed tensor $D$ is then fed into a GRU to capture the dynamic interactions between the target vehicle and its neighbors. The output from the aggregation layer is multiplied by the visual field matrix to simulate changes in focus. This results in a higher dimensional tensor, $ H_{\text{conv}} $, which is stored in $\mathbb{R}^{T \times \text{nbrs}} $. This output allows the model to capture a richer spatial context around the target vehicle in mixed-autonomy traffic environments.

\subsection{Vision-aware Module}
Traditional research often assumes that a driver's attention is evenly distributed throughout the environment [5]. However, it is important to recognize that human attention is limited. Traffic behavior studies have shown that drivers adjust their visual focus based on speed and other factors [6,14]. Specifically, a driver's central visual field (the frontal sector) of attention can be described as a "visual sector". For example, at relatively lower speeds, the driver's focus is concentrated within a visual sector defined by $ r = 30 $ meters and $ \theta = 90^\circ $. In contrast, at higher speeds, the focus shifts to a visual sector with $ r = 90 $ meters and $ \theta = 45^\circ $. This adaptation reflects a narrower but more distant field of view at higher speeds.

Given these findings, our model incorporates an adaptive visual sector concept to simulate the driver's changing visual field at different speeds. Based on the effectiveness confirmed by relevant research [7,14], we define speed thresholds at 30 $km/h$, 60 $km/h$, and 90 $km/h$ to delineate different visual sectors. Different visual sectors are used to apply different weights to these regions. We use a visual matrix $V$ to model the changes in the central visual field. We then perform element-wise multiplication between the interaction-aware output and the visual matrix:
\begin{equation}
N = H_{\text{conv}} \odot V,
\end{equation}
where $V$ is the visual matrix composed of varying values between 0 and 1 depending on the driver's speed. $N$ denotes the node vectors to be fed into the graph neural network.

In mixed-autonomy traffic environments, it is crucial to recognize the different interaction patterns among vehicles in the scene, especially when not all vehicles share the same control logic, including a significant number of human drivers. To address this issue, we utilized two layers of Graph Attention Network (GAT) with ELU activation functions, alternating with dropout layers, to analyze the interactions between targets. This approach enables the simulation of various detection patterns of drivers towards different vehicles in mixed scenarios.

Considering the safety of vehicles in close proximity, we have defined a circular region with a certain safety distance as the " nearby area ". Within this boundary, we enhanced the weights of the GAT adjacency edges. At the same time, we construct a graph, denoted $A^{edge}$, containing all vehicles within the detection area. We compute similarity coefficients between each vertex $N_i$ and its neighboring vehicles. Formally,
\begin{equation}
\alpha(i, j)=\frac{\operatorname{LeakyReLU}\left(\mathbf{a}^{\boldsymbol{T}}\left[W * n_t(i) \| W * n_t(j)\right]\right)}{\sum_{k \in N_i} \operatorname{LeakyReLU} \left(\mathbf{a}^{\boldsymbol{T}}\left[W * n_t(i) \| W * n_t(k)\right]\right)},
\end{equation}
where $n$ denotes the features of a single node. To obtain the final features of each node at time $t$, we sum all the vectors connected to each node:
\begin{equation}
    \hat{n}_t(i)=\sigma\left(\sum_{j \in N_i} \alpha(i, j) W * n_t(j)\right),
\end{equation}
where $\sigma$ is the nonlinear function ELU.

By incorporating the Vision-aware Module, we have not only improved the accuracy of the model but also enabled it to recognize the different driving patterns in mixed-autonomy traffic environments. The output of the module is called the visual feature.

\subsection{Priority-aware Module}
The Priority-aware Module uses a Transformer-based sequence-to-sequence model with an encoder-decoder architecture. Inspired by Transformer translation models, we use the visual feature as input for the encoder and the context feature as input for the decoder. We then perform multi-trajectory prediction based on different behavioral patterns and derive the final trajectories by mapping the results to a bivariate Gaussian distribution.
\subsubsection{Encoder}
The input tensor ${Z}$ to the encoder is a visual feature that can be represented as
${Z} \in \mathbb{R}^{nbrs \times dim},$.
where $nbrs$ is the number of spatial details (e.g. position coordinates) in each trajectory, and $dim$ is the dimension of the feature representation.

The encoder consists of $L$ layers and uses multi-head attention to process three inputs: query (${q}$), key (${k}$), and value (${v}$), all derived from the same input features ${Z}_{{n}}(t)$. Then, a dot product operation followed by a softmax operation is performed between ${Q}$ and ${K}$ to generate pairwise importance weights ${A}$. The weighted sum is then computed using the attention matrix ${A}$ and the value ${V}$:
${Z}^{\prime}={A V}.$
We define the attention function as ${Z}^{\prime}=\operatorname{Attn}({X})$. The multi-head attention mechanism in the encoder module is critical for capturing relevant spatial details. The results of multi-head attention are integrated using a linear transformation with weights ${W}_o \in \mathbb{R}^{n_h h_v \times h_v}$, where $h_v$ is the dimension of the value ${V}$ and $n_h$ is the number of heads:
\begin{equation}
{Z}^{\prime \prime}=\operatorname{MHA}({Z})=\left[\operatorname{Attn}_1({Z}), \ldots, \operatorname{Attn}_{n_h}({Z})\right] {W}_o
\end{equation}
The feedforward module is a 2-layer MLP with activation $\delta(\cdot)$:
\begin{equation}
{Z}^{\prime \prime \prime}=\operatorname{FFN}\left({Z}^{\prime \prime}\right)=\delta\left({Z} {W}_1+{b}_1\right) {W}_2+{b}_2
\end{equation}

\subsubsection{Decoder}\label{Decoder}
The Decoder module of the module receives two input sources: the encoded output ${Z}^{\prime \prime \prime}$ from the encoder and the context feature ${H}_n(t)$ formed in Context-aware Module. The input trajectory tensor ${H}_n(t)$ is embedded using fully connected layers. The Transformer Decoder consists of $L$ layers with multi-head self-attention and position-wise feedforward neural networks. The output of the Transformer denoted as $\tilde{H}^{0} \in \mathbb{R}^{T \times dim}$, is generated using a linear transformation with Dropout and GLU:
\begin{equation}
\tilde{H}^{0}=\left[\tilde{h}_{1}^{0}, \tilde{h}_{2}^{0}, \cdots, \tilde{h}_{T}^{0}\right]
\end{equation}

The actual trajectory of a vehicle in real traffic scenes is often uncertain due to the complexity and diversity of possible driving maneuvers. To address this issue, we introduce the Generator, which is responsible for predicting the probability distribution of different driving maneuvers for the target vehicle[19]. 

Specifically, $\tilde{H}^{0}$ is transformed from historical data into future data using MLP layers, yielding the intention-specific feature $\tilde{Z} \in \mathbb{R}^{F \times dim}$, where $F$ denotes the number of time steps the model is going to predict in the future. In order to predict the probability distribution of the future trajectory at each $t$ time step based on different maneuvers, the intention-specific feature $\tilde{Z}$ and the prediction probability of maneuver classes $P(l a)$ and $P(l o)$are concatenated and then fed into a fully connected layer with weight $W_{m}$ as follows:
\begin{equation}
 e_{t}^{la,lo} = \left(\tilde{Z} \oplus P(la) \oplus P(lo), W_{m}\right)_{\text{MLP}}
\end{equation}

Additionally, to ensure temporal continuity in the predicted trajectory, the encoded representation $e_{t}^{l a, l o}$ at each timestamp is sequentially fed into a decoder Long Short-Term Memory (LSTM) network. The LSTM outputs a bivariate Gaussian distribution for the vehicle's motion, represented as $\theta_{t}$:
\begin{equation}
\theta_{t}=\left(\operatorname{LSTM}\left(e_{t}^{l a, l o}, W_{decoder}\right), W_{d}\right)_{\text{MLP}}
\end{equation}

To estimate the probability distribution of the trajectory prediction $Y$, we apply the total probability theorem to incorporate the uncertainty captured by the bivariate Gaussian distribution. The parameters $\theta$ represent a collection of bivariate Gaussian distribution parameters for each time step in the future trajectory, denoted as $\theta=\left[\theta_{T+1}, \theta_{T+2}, \cdots, \theta_{T+F}\right]$. Each $\theta_{t}$ for $T+1 \leq t \leq T+F$ consists of the mean $\mu_{t, x}$ and $\mu_{t, y}$, the variances $\sigma_{t, x}$ and $\sigma_{t, y}$, and the correlation coefficient $\rho_{t}$ for the vehicle's $\mathrm{x}$ and $\mathrm{y}$ positions. These parameters collectively represent the probability distribution of the predicted trajectory and account for the uncertainty inherent in the prediction process.

\section{Experiments }
In this study, Next Generation Simulation (NGSIM), Highway Drone (HighD), and Macao Connected Autonomous Driving (MoCAD) datasets are used for evaluation. Specifically, three seconds of trajectory history data is used for input, followed by five seconds for prediction. These datasets offer valuable insights into real traffic scenarios, enabling us to draw meaningful conclusions.
\subsection{Experimental Results}
To verify the effect of the model proposed in this paper, we have compared the effect of existing reproducible models, and the results are shown in Table \ref{Table1}.

\begin{table}[!htb]
  \centering
    \caption{Evaluation results for GaVa and the baselines in the three datasets. Note: RMSE (m) is the evaluation metric, where lower values indicate better performance, with some not specifying ('-'). Values in \textbf{bold} represent the best performance in each category.}\label{Table1}
     \setlength{\tabcolsep}{2.5mm}
   \resizebox{\linewidth}{!}{
    \begin{tabular}{c|cccccc}
    \toprule
    \multicolumn{1}{c}{\multirow{2}[2]{*}{Dataset}} & \multirow{2}[3]{*}{Model} & \multicolumn{5}{c}{Prediction Horizon (s)} \\
\cmidrule{3-7}    \multicolumn{1}{c}{} &       & 1     & 2     & 3     & 4     & 5 \\
    \midrule
    \multirow{12}[21]{*}{NGSIM} & S-LSTM [24] & 0.65  & 1.31  & 2.16  & 3.25  & 4.55  \\
          & S-GAN [25] & 0.57  & 1.32  & 2.22  & 3.26  & 4.40  \\
          & CS-LSTM [13] & 0.61  & 1.27  & 2.09  & 3.10  & 4.37  \\
          & MATF-GAN [26]& 0.66  & 1.34  & 2.08  & 2.97  & 4.13  \\
          & NLS-LSTM [27]& 0.56  & 1.22  & 2.02  & 3.03  & 4.30  \\
          &M-LSTM [28]& 0.58  & 1.26  & 2.12  & 3.24  & 4.66 \\
          &IMM-KF [29]& 0.58  & 1.36  & 2.28  & 3.37  & 4.55 \\
          &GAIL-GRU [30]& 0.69  & 1.51  & 2.55  & 3.65  & 4.71 \\
          &MFP [31]& 0.54  & 1.16  & 1.89  & 2.75  & 3.78  \\
          &DRBP[32]& 1.18  & 2.83  & 4.22  & 5.82  & - \\
          & DN-IRL [34]& 0.54  & 1.02  & 1.91  & 2.43  & 3.76  \\
          & WSiP [8]& 0.56  & 1.23  & 2.05  & 3.08  & 4.34  \\
          & CF-LSTM [35]& 0.55  & 1.10  & 1.78  & 2.73  & 3.82  \\
          & MHA-LSTM [33]& 0.41  & 1.01  & 1.74  & 2.67  & 3.83  \\
          & HMNet [36]& 0.50  & 1.13  & 1.89  & 2.85  & 4.04  \\
          & TS-GAN [37]& 0.60  & 1.24  & 1.95  & 2.78  & 3.72  \\
          & STDAN [19]& \textbf{0.39}  & 0.96  & 1.61  & 2.56 & 3.67  \\
          & \textbf{GaVa} & 0.40 & \textbf{0.94} & \textbf{1.52} & \textbf{2.24} & \textbf{3.13} \\ 
    \midrule
    \multirow{7}[18]{*}{HighD} &S-LSTM [24]& 0.22  & 0.62  & 1.27  & 2.15  & 3.41  \\
    &S-GAN [25]& 0.30  & 0.78  & 1.46  & 2.34  & 3.41  \\
    &WSiP [8]& 0.20  & 0.60  & 1.21  & 2.07  & 3.14  \\
    &CS-LSTM(M) [13]& 0.23  & 0.65  & 1.29  & 2.18  & 3.37  \\
    &CS-LSTM [13]& 0.22  & 0.61  & 1.24  & 2.10  & 3.27  \\
    &MHA-LSTM [33]& 0.19  & 0.55  & 1.10  & 1.84  & 2.78  \\
    &MHA-LSTM(+f) [33]& 0.06  & 0.09  & 0.24  & 0.59  & 1.18  \\
    &NLS-LSTM [27]& 0.20  & 0.57  & 1.14  & 1.90  & 2.91  \\
    &DRBP[32]& 0.41  & 0.79  & 1.11  & 1.40  & - \\
    &EA-Net [38] & \textbf{0.15}  & 0.26  & 0.43  & 0.78  & 1.32  \\
    &CF-LSTM [35]& 0.18  & 0.42  & 1.07  & 1.72  & 2.44  \\
    &STDAN [19]& 0.19  & 0.27  & 0.48  & 0.91  & 1.66  \\
    & \textbf{GaVa} & 0.17 & \textbf{0.24} & \textbf{0.42} & \textbf{0.86} & \textbf{1.31} \\
     \midrule
    
    \multirow{7}[10]{*}{MoCAD}
    &S-LSTM [24] & 1.73  & 2.46  & 3.39  & 4.01  & 4.93 \\
    &S-GAN [25] & 1.69  & 2.25  & 3.30  & 3.89  & 4.69  \\
    &CS-LSTM(M) [13]& 1.49  & 2.07  & 3.02  & 3.62  & 4.53   \\
    &CS-LSTM [13] & 1.45  & 1.98  & 2.94  & 3.56  & 4.49  \\
    &MHA-LSTM [33] & 1.25  & 1.48  & 2.57  & 3.22  & 4.20  \\
    &MHA-LSTM(+f) [33] & 1.05  & 1.39  & 2.48  & 3.11  & 4.12  \\
    &NLS-LSTM [27] & 0.96  & 1.27  & 2.08  & 2.86  & 3.93\\
    &WSiP [8] & 0.70  & 0.87  & 1.70  & 2.56  & 3.47  \\
    &CF-LSTM [35] & 0.72  & 0.91  & 1.73  & 2.59  & 3.44 \\
    &STDAN [19] & 0.62  & 0.85  & 1.62  & 2.51  & 3.32  \\
    & \textbf{GaVa} & \textbf{0.51} & \textbf{0.75} & \textbf{1.34} & \textbf{2.13} & \textbf{2.91} \\
    \bottomrule
    \end{tabular}%
  \label{tab:addlabel}%
  }
\end{table}%
The accuracy of the proposed model in this paper was evaluated by comparing it to the aforementioned baseline using Root Mean Square Error (RMSE). The analysis indicates that the model consistently produces more accurate results across different time dimensions, regardless of the dataset used. 

Previous research has demonstrated that incorporating information from surrounding vehicles leads to superior results compared to not utilizing such relevant information. This paper compares all baselines based on interactions between the ego vehicle and other vehicles. Among the reproducible models selected in this study, we found that the S-LSTM, which solely employs fully connected layers for prediction, and the trajectory prediction model based on GAN generation performed the worst. The social convolutional networks and wave theory model using amplitude and phase yielded better results. Prior to this study, the model based solely on the multi-head attention mechanism achieved the best performance. 

From the perspective of traffic behavior, we constructed the interaction part of the model and simulated the driver's visual focus changing with vehicle speed using a multi-head transformer model, which performed best. In the 5-second prediction phases of the three data sets, it outperformed the former model by 15.2\%, 19.4\%, and 12.0\%, respectively.

Based on the experimental results, two conclusions can be
drawn: 1) Models that use multimodal trajectories for loss propagation are superior because they consider multiple different states of the model predictions and determine the optimal solution through a comprehensive evaluation.
2) Convolutional networks are more effective compared to simulating physical factors, possibly because the computational approach of computers is better suited to convolutional spatial operations.
This experiment demonstrates that the combination of convolutional networks and multi-head attention mechanisms can achieve excellent results. This is likely due to their cooperation, which increases the model's robustness.

\subsection{Qualitative Results}
In addition, we performed simulations to visualize predicted vehicle trajectories over a 5-second interval based on the experimental NGSIM and HighD datasets. As shown in Figure 3, the visual representation vividly illustrates the predictive capabilities of multimodal trajectories (shown as green solid lines) in complex traffic scenarios.

\subsection{Ablation Studies} \label{Ablation Studies}
In this section, we perform an ablation study of the model to demonstrate the necessity and conciseness of the inter-modularity of the GaVa model. Specifically, the model was altered into the following variants:
\begin{enumerate}
\item GVTA (-IaM): This variant directly removes the entire Interaction-aware module, i.e., it allows the model to make predictions relying solely on each vehicle's independent historical data without accessing the interaction characteristics of all vehicles through the graph neural network.
\item GVTA (-VM): This variant removes visual matrix, i.e., ignores the idea that the focus of the field of view varies with speed, and removes the corresponding module in the Vision-aware Module.
\item GVTA (+NVM): This variant uses not only the results of the Vision-aware Module outputs as node feature vectors but also adds the Interaction-aware Module results that have not been recognized by the visual matrix as node feature vectors.
\end{enumerate}
\begin{table}[!ht]
    \centering
    \caption{Prediction Error of Ablation Models on NGSIM datasets}
    \label{Table2}
    \begin{tabular}{ccccccc}
    \hline
        Time (s) & GaVa (-IaM) & GaVa (-VaM) & GaVa (+NVM) & GaVa  \\ \hline
        1 & 0.62 & 0.45 & 0.43 & $\mathbf{0.39}$ \\ 
        2 & 1.43 & 1.16 & 0.96 & $\mathbf{0.93}$ \\ 
        3 & 2.45 & 2.13 & 1.59 & $\mathbf{1.52}$  \\ 
        4 & 3.89 & 3.35 & 2.37 & $\mathbf{2.24}$  \\ 
        5 & 5.14 & 4.26 & 3.36 & $\mathbf{3.13 }$ \\ \hline
    \end{tabular}
\end{table}
We performed ablation experiments on the above three variants with the same set of other model parameters and training hyperparameters on NGSIM datasets, and the results are shown in Table \ref{Table2}. From the results, it is not difficult to draw the following conclusions: First, the variant that ignores the composition of social network relationships performs the weakest, even worse than S-LSTM, confirming the importance of capturing social network interactions for trajectory prediction. Second, we find that the model variant that removes the visual field is not ideal, and its effect is comparable to that of S-GAN, suggesting that the visual field is critical to the overall role of the model. Third, instead of getting better, the results of the model with added convolutional data not detected by the visual field got relatively worse, with its RMSE increasing more and more as the number of seconds of prediction increased. This suggests that the necessary conciseness of the feature vectors of the graph neural network is beneficial for the final trajectory prediction.

The results confirm that extracting information about inter-vehicle interactions through graph neural networks is necessary for the machine, and in particular, Vision-aware Recognition validates the idea that the visual field of interest region of traffic behavior changes with speed. Both are essential for improving model performance.
\begin{figure}[t]
  \centering  \includegraphics[width=0.85\linewidth]{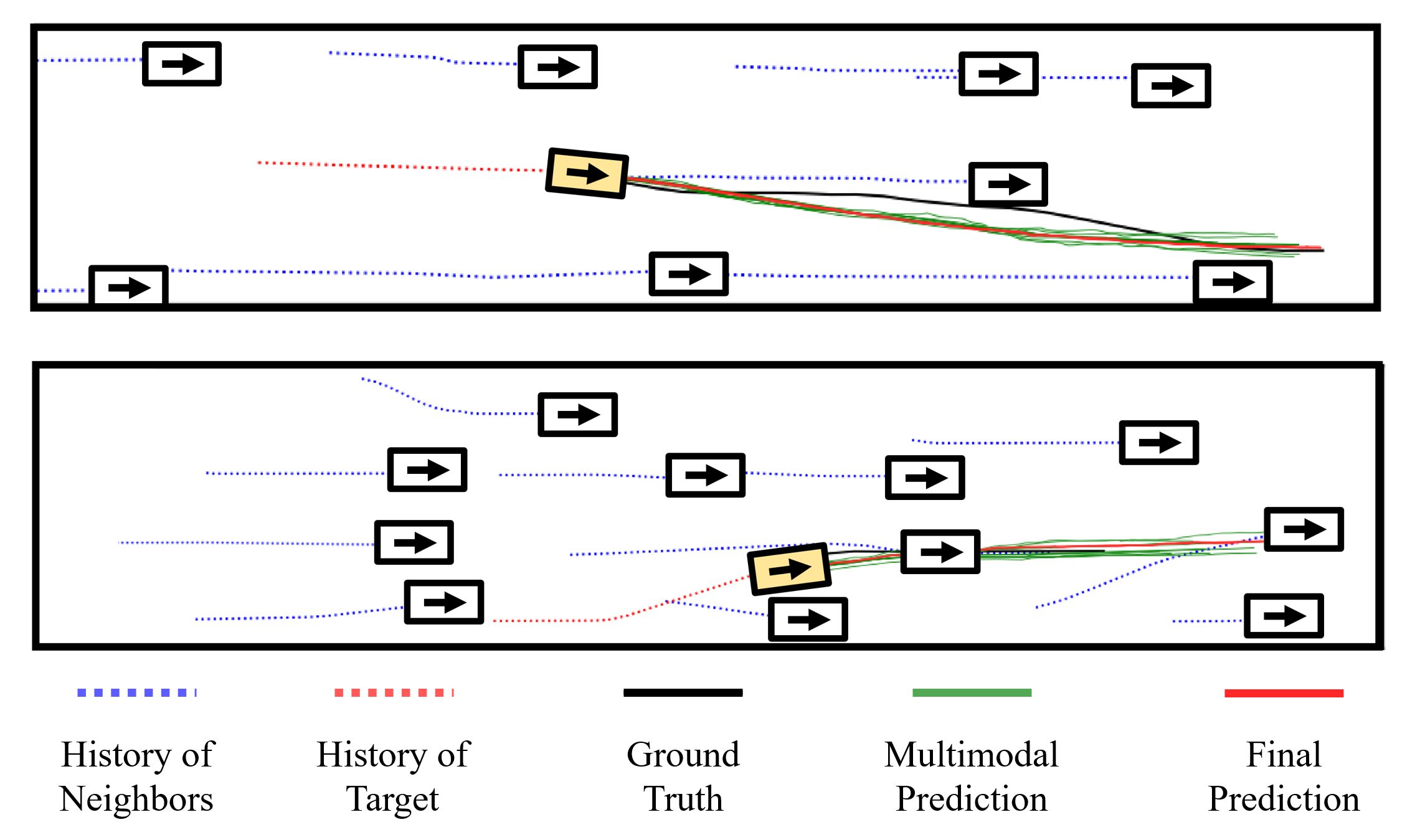} 
  \caption{Visualization of Predictions for Complex Traffic Situations in NGSIM (above) and HighD (below) Datasets. The yellow rectangles denote the target vehicle, while other rectangles represent the neighboring vehicles.}
  \label{fig2}
\end{figure}

\section{Conclusion}\label{Conclusion}
In this study, we have introduced a comprehensive trajectory prediction model called GaVa, which incorporates insights from traffic behavior studies. By incorporating temporal data, spatial data, and visual matrix, GaVa excels in accurately predicting future trajectories. The use of a multi-head attention mechanism and a novel graph neural network composition contribute to GaVa's superior trajectory prediction performance on three datasets, outperforming reproducible state-of-the-art results. Through an ablation study, we validate the importance of each component in the model and emphasize the need to integrate traffic behavioral science. Our work demonstrates the potential of combining domain knowledge from traffic behavior studies with advanced neural network architectures. We believe that this fusion of expertise opens promising avenues for future research in trajectory prediction and autonomous driving.

\end{document}